\documentclass[journal]{elsarticle}
\usepackage{amsmath,graphicx}
\usepackage{amsmath,amssymb} 
\usepackage{epsfig}
\usepackage{multirow,color}

\begin{document}


\begin{frontmatter}

%
%

%
%
\title{Progressive Generative Adversarial Networks for Medical Image Super resolution}

%



\author{Dwarikanath Mahapatra$^{1}$, Behzad Bozorgtabar$^{2}$ \\
$^{1}$IBM Research - Australia, $^{2}$EPF Laussane, Switzerland}



\begin{abstract}
Anatomical landmark segmentation and pathology
localisation are important steps in automated analysis of medical
images. They are particularly challenging when the
anatomy or pathology is small, as in retinal images (e.g.
vasculature branches or microaneurysm lesions) and cardiac MRI, or when the
image is of low quality due to device acquisition parameters as in magnetic resonance (MR) scanners. We propose an image super-resolution method using
progressive generative adversarial networks (P-GANs) that can
take as input a low-resolution image and generate a high
resolution image of desired scaling factor. The super resolved
images can be used for more accurate detection of landmarks
and pathologies. Our primary contribution is in proposing a
multi stage model where the output image quality of one stage
is progressively improved in the next stage by using a triplet
loss function. The triplet loss enables stepwise image quality
improvement by using the output of the previous stage as the
baseline. This facilitates generation of super resolved images
of high scaling factor while maintaining good image quality.
Experimental results for image super-resolution show that our
proposed multi stage P-GAN outperforms competing methods
and baseline GANs. The super resolved images when used
for landmark and pathology detection result in accuracy levels close to those obtained when using the original high resolution images. We also demonstrate our method’s effectiveness on magnetic resonance (MR) images, thus
establishing its broader applicability 
\end{abstract}



\end{frontmatter}


\section{Introduction}
\label{sec:intro}

Retinal fundus image analysis is essential for diagnosis of retinal conditions such as glacuoma and diabetic retinopathy. An important component of automated diagnosis of retinal conditions is the detection of pathologies (haemorrhages, microaneurysms, exudates) and landmarks (vasculature, optic cup and disc, fovea). Fundus image resolution is high enough to detect and segment prominent landmarks (e.g., optic disc, fovea, main vessels) and pathologies (e.g., hard exudates). However there are many pathologies which cover a very small area in the fundus images (e.g.,  microaneurysms, haemorrhages) or are not clearly visible (e.g. soft exudates and certain neovascularizations). Smaller branches of the vasculature are difficult to segment in normal fundus images and hence it is desirable to have higher resolution local image patches covering the specific pathologies to facilitate detailed disease analysis. We propose a image super-resolution algorithm using progressive generative adversarial networks (P-GANs) that takes as input a low-resolution image patch and outputs a high-resolution image to facilitate more accurate diagnosis. Our method's effectiveness is demonstrated using retinal vasculature segmentation and microaneurysm detection. We also show results on magnetic resonance (MR) images for cardiac left ventricle segmentation, thus establishing the wider applicability of our method.

\subsection{Related Work}

Interpolation of medical images (such as MRI) leads to partial volume effects that affect the final segmentation. There exist many methods using super-resolution (SR) on medical images \cite{Jog7,Jog8}. A common approach to image super-resolution (ISR) are example based methods \cite{Jog8} which leverage the information from high-resolution (HR) images together with a low-resolution (LR) image and generate an SR version approximating the original HR image. Self similarity \cite{Jog7} and generative models \cite{Jog5} have also been used for SR. These methods are too reliant on external data which may not always be available, thus putting them at an disadvantage. 

Single image based SR methods downsample a given image to create a LR image and learn the mapping between the original and LR version. The learnt mapping is then applied to the original image to generate a SR image. In \cite{Jog9} HR and LR dictionaries are learned from MRI to generate SR images. These methods depend on learning the dictionaries on external LR-HR images and assume that the test image is a representative of the training data. Since this is not always the case the results are unsatisfactory. 
%
%
%
These approaches are computationally demanding as the candidate patches have
to be searched in the training dataset to find the most suitable HR candidate.
Instead, compact and generative models can be learned from the training data
to define the mapping between LR and HR patches.

Parametric generative models,
such as coupled-dictionary learning based approaches, have been proposed
to upscale MR brain \cite{Jog9} and cardiac \cite{Oktay3} images. These methods benefit from sparsity constraint to express the link between LR and HR. 
Similarly, random  forest based non-linear regressors have been proposed to predict HR patches from LR data and have been successfully applied on diffusion tensor images
\cite{TannoMICCAI16}. Recently, convolutional neural network (CNN) models \cite{Oktay5} have been put
forward to replace the inference step as they have enough capacity to perform
complex nonlinear regression tasks. Even by using a shallow network composed
of a few layers, \cite{Oktay5} achieve superior results over other state-of-the art SR methods.
Recent works have proposed image SR methods based on training data free approach using Fourier burst accumulation \cite{Jog}, CNNs \cite{Oktay} and generative adversarial  networks (GANs) \cite{Srgan,Mahapatra_MICCAI17}.

\subsection{Our Contribution}

In an earlier work \cite{Mahapatra_MICCAI17} we proposed a GAN based image SR method that incorporated saliency maps. The saliency map based approach had some limitations such as: 1) choice of optimal window size and weights for saliency map calculation depended on the specific image and was non-optimal; 2) consequently, the super-resolution output was unsatisfactory for certain cases.
To overcome these limitations and to avoid a heuristic approach, in this paper we propose a novel image SR method based on multi-stage GANs that uses a triplet loss function. We call our method progressive GAN (P-GAN) since it leverages the triplet loss function to progressively improve the quality of super resolved images. Our current method is different from \cite{Mahapatra_MICCAI17} since: 1) it uses progressive GANs and triplet loss function; 2) it does not use any heuristic based parameter values; and 3) it outperforms \cite{Mahapatra_MICCAI17} for different scaling factors.

GANs \cite{Srgan21} are used to learn a generative model of images that is similar to a given set of training images. They have been used in various applications such as image super resolution \cite{Srgan}, image registration \cite{MahapatraGAN_ISBI18}, active learning \cite{MahapatraAL_MICCAI18}, image synthesis and image translation using conditional GANs (cGANs) \cite{CondGANs} and cyclic GANs (cycleGANs) \cite{CyclicGANs}.  Multiple blocks of deep residual network (ResNet) \cite{Srgan27} with skip connections are used to construct the generator network. In our proposed multi-stage approach we use the output of one stage as the input to the next stage. The image generation framework in the next stage uses triplet loss to improve the quality of the image from the previous stage. This ensures that good quality images are generated for high scaling factors. Combining the generator network with a discriminator encourages solutions that preserve the information content and perceptual information of an image. This leads to HR images that do not compromise on perceptual clarity and result in better retinal vasculature segmentation and microaneurysm (MA) detection results, as well as MRI organ segmentation.

Our paper makes the following contributions: 1) a novel P-GAN architecture using multiple stages of GANs is proposed that enables generation of SR images of high scale factors (upto $32$); 2) While GAN based methods use conventional mean square error (MSE) and CNN feature loss values to generate SR images, we use an additional triplet loss function to improve image quality from one stage to the next. This ensures that image quality is not compromised despite generating images of high scale factors. A combination of multi stage GANs and triplet loss helps us outperform conventional GAN based methods \cite{Srgan}.
The rest of the paper is organized as follows: Section~\ref{sec:met} describes basics of GANs and our novel contribution, followed by experimental results in Section~\ref{sec:expt} and our conclusion in Section~\ref{sec:concl}.

\section{Methodology}
\label{sec:met}

In this section we will first give a brief outline of GANs (Section~\ref{met:GANs}) and then explain our contribution in the form of progressive-GANs (Section~\ref{met:pGANs})

\subsection{Generative Adversarial Networks}
\label{met:GANs}

\begin{figure}[t]
\begin{tabular}{c}
\includegraphics[height=5.2cm, width=9.4cm]{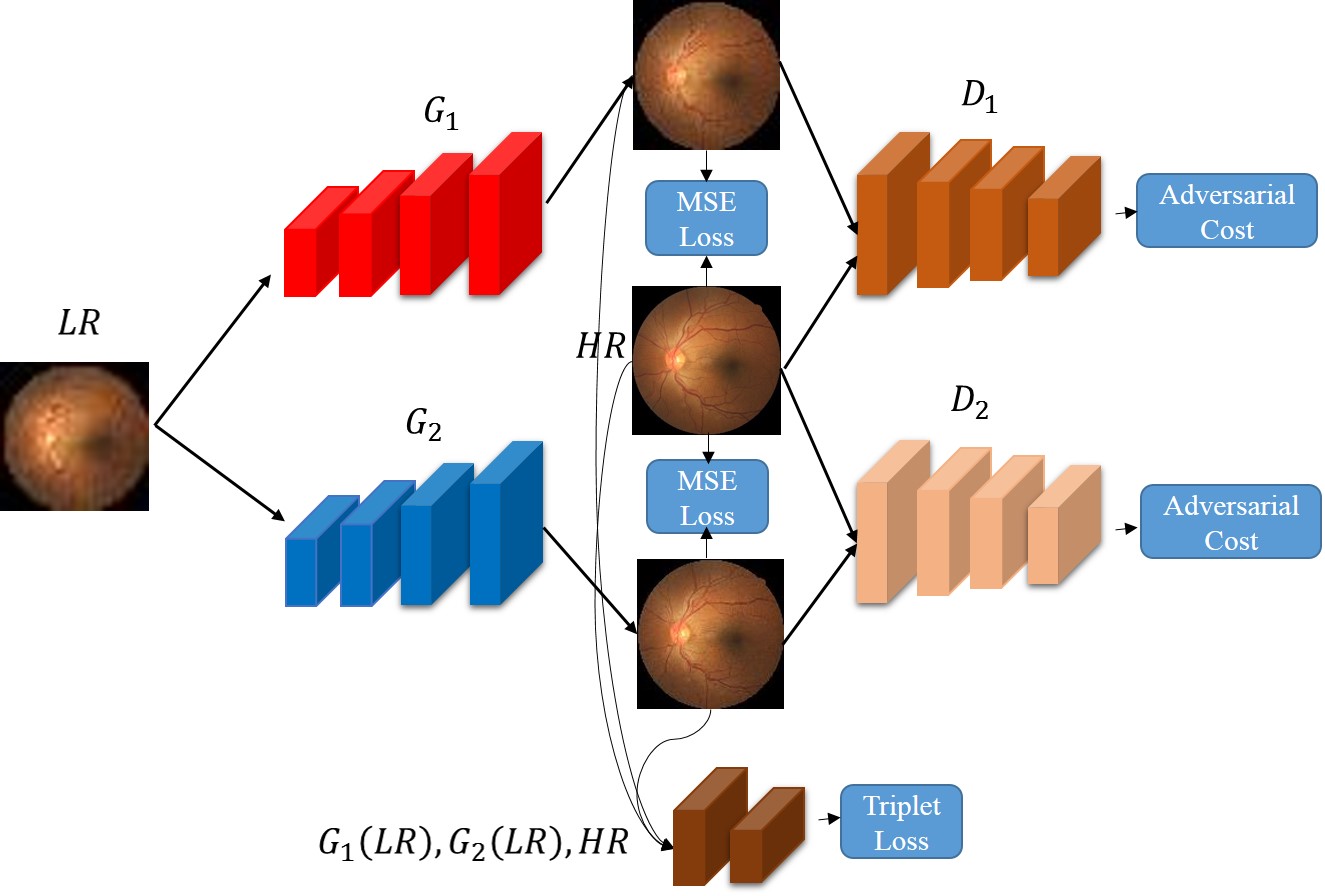}  \\
\end{tabular}
\caption{Depiction of Progressive GAN architecture. }
\label{fig:PGan1}
\end{figure}

\begin{figure}[t]
\begin{tabular}{c}
\includegraphics[height=4.2cm, width=9.9cm]{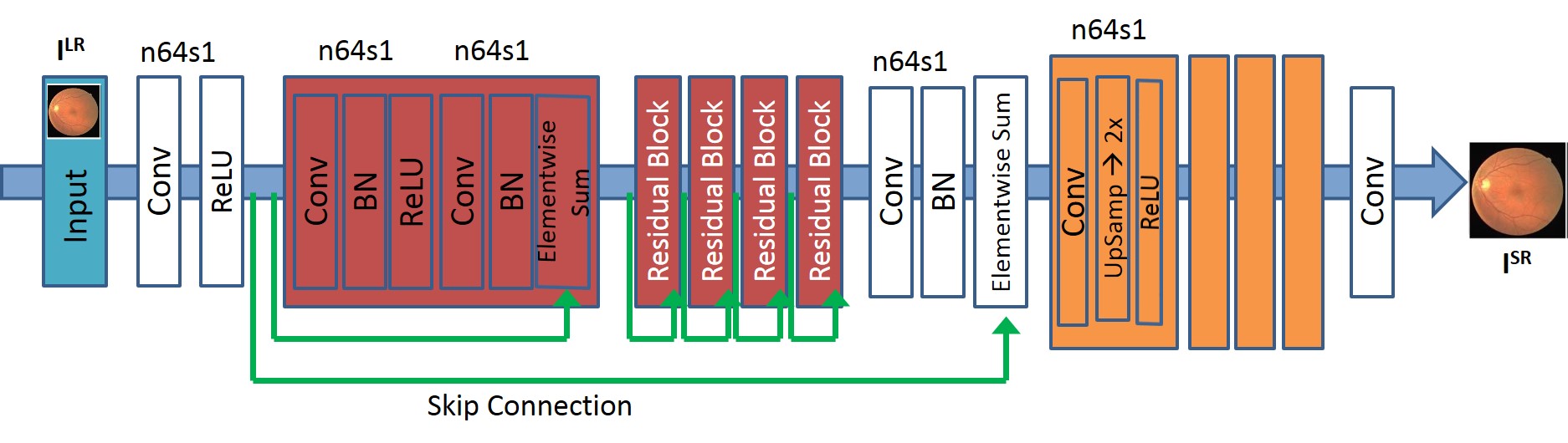}  \\
(a) \\
\includegraphics[height=4.2cm, width=9.9cm]{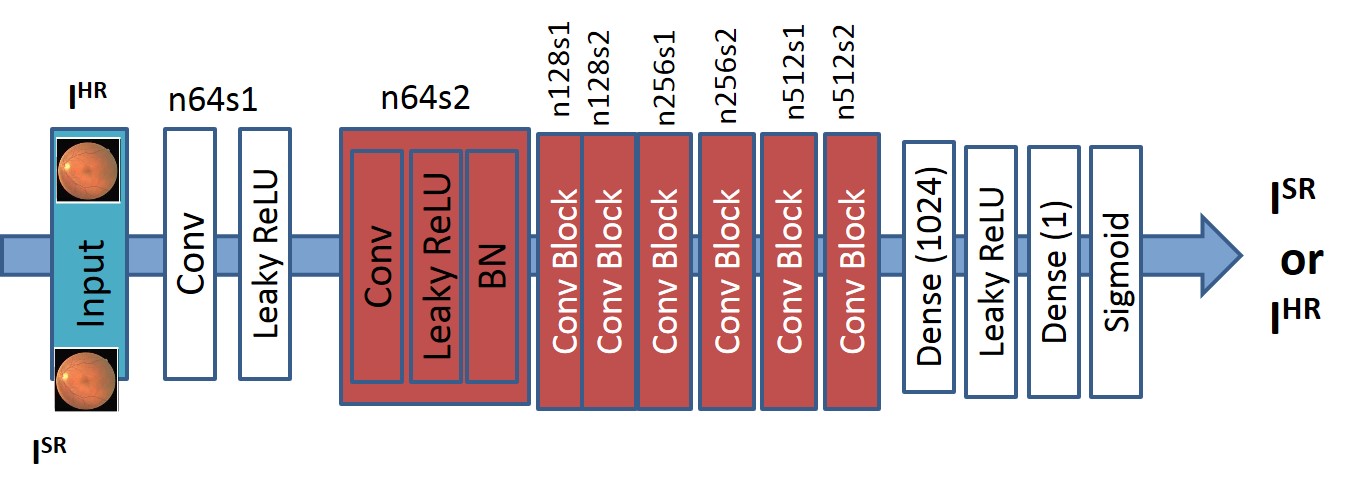}  \\
(b)\\
\end{tabular}
\caption{(a) Generator Network; (b) Discriminator network. $n64s1$ denotes $64$ feature maps (n) and stride (s) $1$ for each convolutional layer. }
\label{fig:Gan}
\end{figure}

Super resolution estimates a high-resolution image $I^{SR}$ from a low-resolution input image $I^{LR}$. Figure~\ref{fig:PGan1} shows the architecture of a progressive GAN setup where the output of the first stage is used as input to the second stage, and the triplet loss is used from the second stage onwards to improve super resolution results. Each super-resolution stage consists of a generator and discriminator network which are depicted in Figure~\ref{fig:Gan}. For training, $I^{LR}$ is the low-resolution version of the high-resolution counterpart $I^{HR}$, obtained by applying a Gaussian filter to $I_{HR}$ followed by downsampling with factor $\textbf{r}$. 
The generator network is a feed-forward CNN, $G_{\theta_G}$, parametrized by $\theta_G = {W;b}$, the weights and biases of a L-layer network. The parameters are obtained by, 
\begin{equation}
\widehat{\theta}=\arg \min_{\theta_G} \frac{1}{N} \sum_{n=1}^{N} l^{SR}\left(G_{\theta_G}(I_n^{LR}),I_n^{HR}\right),
\label{eq:theta1}
\end{equation}
where $l^{SR}$  is the loss function and $I_n^{HR}$, $I_n^{LR}$ are the set of high-resolution and low-resolution images.
In the generator (Figure~\ref{fig:Gan} (a)) the input image $I^{LR}$ is passed through a convolution block followed by ReLU activation. The output is passed through a residual block with skip connections. Each block has convolutional layers with $3\times3$ filters and $64$ feature maps,  followed by batch normalization and ReLU activation. This output is subsequently passed through multiple residual blocks. Their output is passed through a series of upsampling stages, where each stage doubles the input image size. The output is passed through a convolution stage to get the super resolved image $I^{SR}$. Depending upon the desired scaling, the number of upsampling stages can be changed.  
%
 The adversarial min-max problem is defined by,
%
%
\begin{equation} 
\min_{\theta_G} \max_{\theta_D}  \mathop{\mathbb{E}}_{I^{HR}~p_{train}(I^{HR})}[\log D_{\theta_D}(I^{HR})]  + \mathop{\mathbb{E}}_{I^{LR}~p_{G}(I^{HR})}[\log (1-D_{\theta_D}(G_{\theta_G}(I^{HR}))] 
\label{eq:cf1}
\end{equation}
This trains  a generative model $G$
 with the goal of fooling a
differentiable discriminator $D$ that is trained to distinguish
super-resolved (SR) images from real images. 
$G$ creates solutions that are very similar to real images and thus difficult to classify by $D$.
This encourages perceptually superior solutions  and is superior
to solutions obtained by minimizing pixel-wise MSE.
%

$D$  solves the maximization problem in Eqn.~\ref{eq:cf1}.
The discriminator network (Figure~\ref{fig:Gan} (b)) has multiple 
convolutional layers with the 
kernels increasing by a factor of $2$ from $64$ to $512$. Leaky ReLU is used and strided convolutions reduce the image dimension  when the number of
features is doubled. The resulting $512$ feature maps are
followed by two dense layers and a final sigmoid activation to obtain a probability map, which is used to classify the image as real or fake.

\subsection{Loss Function}

 $l^{SR}$ is a combination of content loss ($l_{Cont}^{SR}$) and adversarial or generative loss ($l_{Gen}^{SR}$), balanced by a factor $\alpha=0.01$, and is given by  : 
\begin{equation}
l^{SR} = l_{Cont}^{SR} + \alpha l_{Gen}^{SR}
\label{eq:cf2}
\end{equation}
%
%
%

\subsubsection{Content Loss - $l_{Cont}^{SR}$}
%
The first component of $l_{Cont}^{SR}$ is the mean square error (MSE) loss ($l_{MSE}$),
%
%
\begin{equation}
l_{MSE}=\frac{1}{WH} \sum _{x=1}^{W} \sum _{y=1}^{H} (I^{HR}_{x,y} -G_{\theta_G} (I^{LR})_{x,y} )^2,
\label{eq:wlMSE}
\end{equation}
where $I^{HR}$ is the high-resolution image and $G_{\theta_G} (I^{LR})$ is the super resolved image.
%
%
Next, a CNN loss \cite{Srgan} is calculated as the $L2$ distance between SR image and ground-truth HR image using all $512$ feature maps of Relu $4-1$ layer of a pre-trained $VGG-16$ \cite{VGG}. It is defined as,
\begin{equation}
l^{SR}_{CNN} =\frac{1}{W_{i,j}H_{i,j}} \sum _{x=1}^{W_{i,j}} \sum _{y=1}^{H_{i,j}} (\phi_{i,j}(I^{HR})_{x,y} -\phi_{i,j}(G_{\theta_G} (I^{LR}))_{x,y} )^2
\label{eq:lCNN}
\end{equation}
$\phi_{i,j}$ the feature map obtained by the $j-$th convolution  (after activation) before the $i-$th max pooling layer and
$W_{i,j}$ and $H_{i,j}$ are the dimensions of $\phi$.



\subsubsection{Adversarial Loss}
The generative (or adversarial) loss $l^{SR}_{Gen}$ over all training samples is 
%
\begin{equation}
l^{SR}_{Gen}=\sum_{n=1}^N -\log D_{\theta_D}(G_{\theta_G}(I^{LR}))
\end{equation}
$D_{\theta_D}(G_{\theta_G}(I^{LR}))$  is probability that  $G_{\theta_G}(I^{LR})$ is a natural HR image. This network favours solutions in the manifold of retinal images. Convergence is facilitated by minimizing $-\log D_{\theta_D}(G_{\theta_G}(I^{LR}))$ instead of $-\log [1-D_{\theta_D}(G_{\theta_G}(I^{LR}))]$.


\subsection{Progressive Generative Adversarial Networks}
\label{met:pGANs}

In the previous sections we have described the working of a conventional GAN with a coupled generator and discriminator. The GAN based approach of \cite{Srgan} does not produce good quality results for scaling factors above $8$. An obvious way to generate higher scale images is to couple successive generator networks or upsampling stages to obtain images of desired scaling factor. The shortcoming of this approach is that any errors of one stage are propagated to the other step as there is no correction mechanism. We propose a multi-stage architecture with a triplet loss based in-built correction mechanism that helps to compensate any quality degradation of images generated st each stage. In this section we explain our novel P-GAN method.

We denote our method as progressive GAN for image super-resolution ($P-SRGAN$).
 $P-SRGAN$ works by first taking a LR image as input which is processed through the first generator network $G1$ to obtain a super resolved image of scale factor $2$.  Similar to a conventional GAN there is a discriminator network $D1$ that helps $G1$ generate good quality images by comparing with high-resolution (HR) images. The loss function combines MSE and CNN loss terms. We denote the SR image from $G1$ as $I^{SR}_{1}$. $I^{SR}_{1}$ is passed through another network $G2$ to generate an image further upsampled by a factor $2$, which we denote as $I^{SR}_{2}$. However, in generating $I^{SR}_{2}$ we make changes to the cost function by including an additional triplet loss  component \cite{TripletLoss}. Note that for each stage we upsample the image by $2$x, which implies that there is only one upsampling block in the generator.

The triplet loss has three variables in it - positive, negative and anchor. The cost function is such that it minimizes the distance between the anchor ($x^{a}$) and positive ($x^{p}$), while maximizing the distance between the anchor and the negative ($x^{n}$). Thus the loss being minimized is,
\begin{equation}
\sum_i^{N} \left[ \left\| f(x^{a}_i) - f(x^{p}_i)\right\|^{2}_2 - \left\| f(x^{a}_i) - f(x^{n}_i)\right\|^{2}_2 \right]
\end{equation}

 In our formulation the anchor is $I_{2}^{SR}$, the negative is $I^{SR}_{1}$ while the positive is the ground truth HR image $I^{HR}$. We would like the generated image at the second (and subsequent) stage, $I^{SR}_{2}$, to be as different as possible from the first (or previous) generated image, $I^{SR}_{1}$. At the same time we enforce that $I^{SR}_{2}$ should be as similar as possible to $I^{HR}$. The rationale behind such an approach is that $I^{SR}_{2}$ should improve upon $I^{SR}_{1}$ and both of them should be as different as possible. At the same time $I^{SR}_{2}$ should be as similar to $I^{HR}$ in order to improve upon the quality of $I^{SR}_{1}$. These dual constraints ensure that $I^{SR}_{2}$ is a quantitative and qualitative improvement over $I_{1}^{SR}$ and closer to $I^{HR}$. For subsequent steps of upsampling by factor $2$ we employ the same loss function where the anchor is the generated image at stage $n$ $I_{n}^{SR}$, the negative is $I_{n-1}^{SR}$ and the positive image is always the ground truth $I^{HR}_n$ for stage $n$. In addition to the triplet loss, MSE and CNN loss are used for all stages $n\geq2$.



\subsection{Training Details:}

We trained all networks on a NVIDIA Tesla $K40$ GPU having $12$ GB RAM.
The LR images were obtained by downsampling the original HR images  with a
bicubic kernel of varying downsampling factors. 
The intensity of all images (fundus and MRI) was scaled to $[0,1]$. 
VGG feature maps were rescaled by a factor of $\frac{1}{12.75}$ (based on actual values of the two feature maps)
to obtain VGG losses of a scale that is comparable to the
MSE loss. 
For optimization we use Adam with $\beta_1 = 0.9$. The SRResNet networks
were trained with a learning rate of $10^{-4}$ for $10^{6}$ update
iterations. We employed the trained MSE-based SRResNet
network as initialization for the generator when training
the actual GAN to avoid undesired local optima. GAN based methods such as ours and \cite{Srgan} are trained with $10^{5}$ update iterations
at a learning rate of $10^{-4}$ and another $10^{5}$ iterations at a
lower rate of $10^{-5}$. During test time
batch-normalization update is off such that the output is 
deterministic and depends on the input.  Our entire pipeline was implemented with Python and TensorFlow was used for the neural network architectures.


\section{Experiments And Results}
\label{sec:expt}

In this section we demonstrate the effectiveness of our super resolution algorithm by reporting results on different kinds of images. We report ISR performance on retinal colour fundus images and magnetic resonance (MR) images. We also show results for segmentation of different landmarks and pathologies in fundus and MR images.

\subsection{Retinal Dataset Description}

To test the effectiveness of our image super resolution algorithm we train it on $5000$ retinal fundus images taken from EYEPACS \cite{kaggle}. The original images were first resized to $1024\times1024$ pixels which was the reference high resolution (HR) image. Subsampled images of scaling factors (r) $\frac{1}{2},\frac{1}{4},\frac{1}{8},\frac{1}{16}$,$\frac{1}{32}$. The high resolution image was then downsampled using a bicubic kernel by a factor of $2$ along the rows and columns. This gave images of size $512\times512$, which are then further downsampled by factors of $2$ to get images of size $256\times256$, $128\times128$, etc. We train our network on the entire set of $5000$ images and use a \emph{separate set} of $1000$ images for testing. 
%
The average training time for a single batch of $5000$ images is $8$ hrs for $r=2$, $15$ hours for $r=4$, $21$ hours for $r=8$, $28$ hours for $r=16$ and $35$ hours for $r=32$. The time taken to generate a super resolved image is $1$ ms for $r=2$, $1.4$ ms for $r=4$, $1.9$ ms for $r=8$, $2.5$ ms for $r=16$ and , $3.3$ ms for $r=32$. The training and test was performed on a NVIDIA Tesla K$40$ GPU with $12$ GB RAM.

\subsection{Image Super Resolution Results}

For comparative analysis, results for super resolution are shown for the following methods:
\begin{enumerate}
\item $P-SRGAN$ - our proposed progressive method  using a multi stage architecture in combination with triplet loss to generate super resolved images of desired scaling factor. Our method's difference from \cite{Srgan} is in the use of a multi stage architecture and triplet loss.

\item $SRGAN_{Sal}$ - the saliency map based SR method of \cite{Mahapatra_MICCAI17}.

\item $P-SRGAN_{MSE}$ - our proposed progressive method with the exception that instead of triplet loss we use MSE loss in every stage. This is designed to highlight the importance of triplet loss in improving quality of super resolved images.

\item $SRGAN_{Ledig}$ - the original GAN based super resolution algorithm of \cite{Srgan}. Based on the original work, the desired scaling factor is obtained by additional pixel upsampling blocks in the original architecture

\item $SRCNN$- the CNN based image super resolution algorithm of \cite{SRCNN}. 

\item $SR-RF$- the random forest based image super resolution method of \cite{SR-RF}.

\item $SSR$- the self super resolution method of \cite{Jog}.
\end{enumerate}

The results of the different methods are compared using the following measures: 1) peak signal to noise ratio (PSNR);  2) structural similarity (SSIM) \cite{SSIM}; and 3) $S3$ - the sharpness metric of \cite{S3Metric}. These metrics are calculated between the generated SR image and ground truth HR image.
Since the original retinal fundus images had different dimensions, they are first resized to $1024\times1024$ since this was the smallest image dimension in the dataset. These images act as the ground truth HR images. They are then downsampled by different scaling factors to generate LR images. These LR images are used to generate SR images of the corresponding scaling factor, and are compared with the ground truth HR images. The quantitative results for scaling factors $4-32$ are presented in Table~\ref{tab:ISR_res1}. 

It is quite obvious that our proposed approach performs the best among all methods, even outperforming $SRGAN_{Sal}$. An interesting observation is that replacing the triplet loss by conventional MSE loss does not result in much improved results compared to $SRGAN_{Ledig}$. This clearly demonstrates that the use of triplet loss is an important contribution in improving the quality of super resolved images.
While upscaling by a factor of $2$ the performance of all methods is similar and hence we do not show the corresponding results. The difference in performance between our approach and other competing methods is noticeable for scaling factor $4$ and to a larger degree for scaling factors above $8$.

\begin{figure*}[t]
\begin{tabular}{cccc}
\includegraphics[height=3.2cm,width=3.2cm]{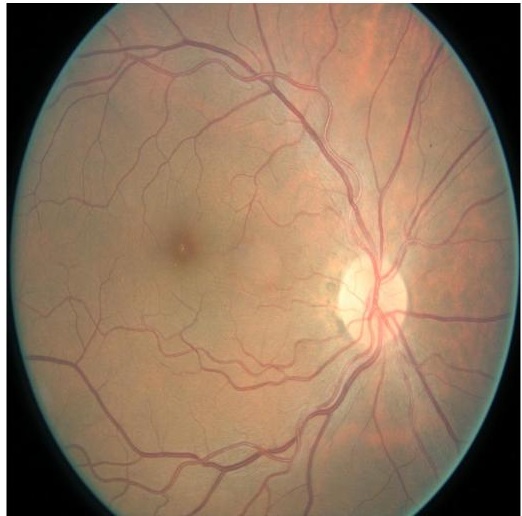} &
\includegraphics[height=3.2cm,width=3.2cm]{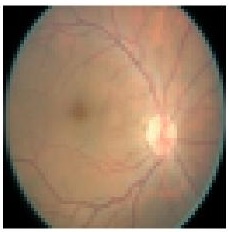} &
\includegraphics[height=3.2cm,width=3.2cm]{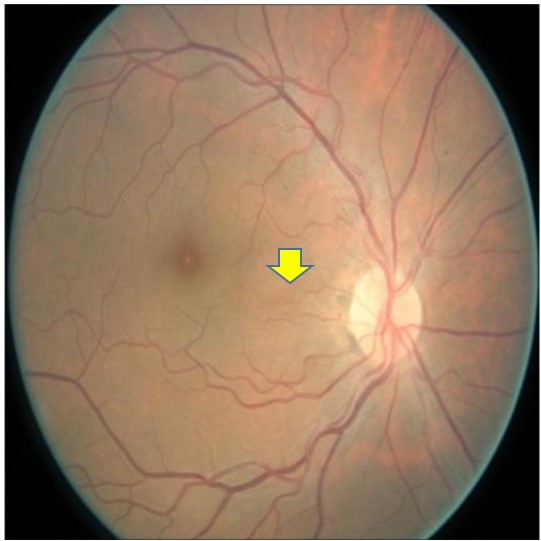} &
\includegraphics[height=3.2cm,width=3.2cm]{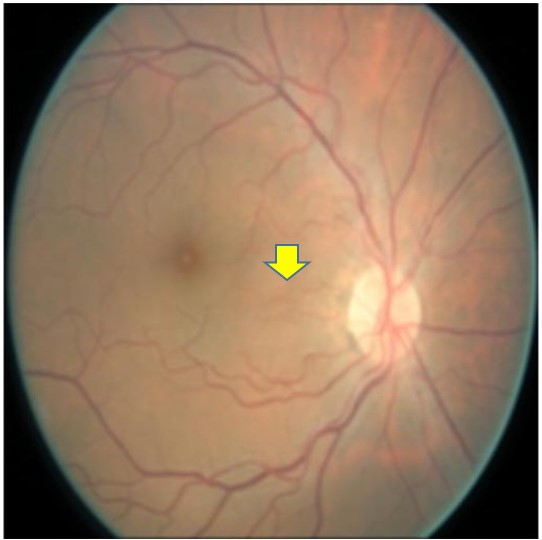} \\
(a) & (b) & (c) & (d) \\
\includegraphics[height=3.2cm,width=3.2cm]{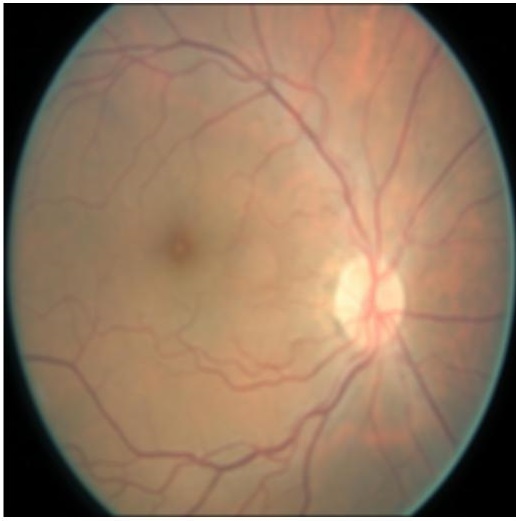} &
\includegraphics[height=3.2cm,width=3.2cm]{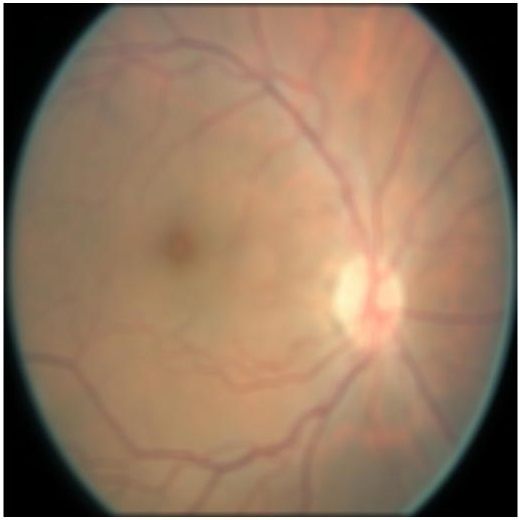} &
\includegraphics[height=3.2cm,width=3.2cm]{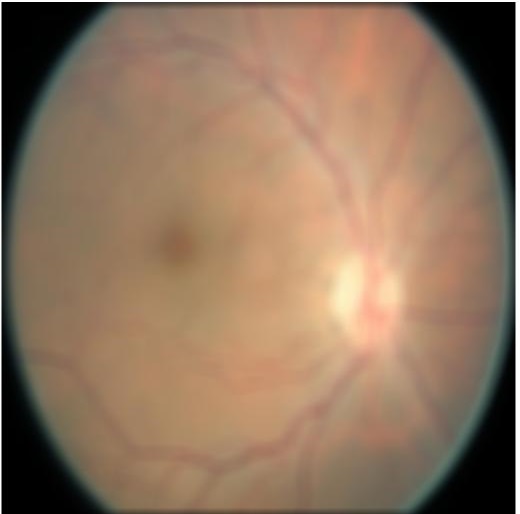} & 
\includegraphics[height=3.2cm,width=3.2cm]{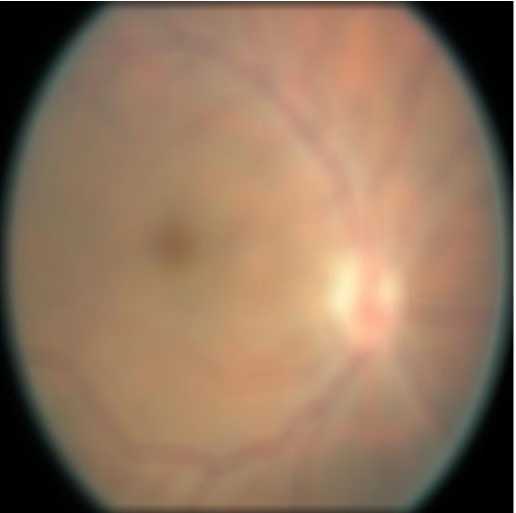}\\

(e) & (f) & (g) & (h)\\ 
\end{tabular}
\caption{Example results for retinal image super resolution.  (a) orginal HR images; (b) Low resolution subsamled by a factor of $16$; super resolved images using (c) $P-SRGAN$; (d) $SRGAN_{Sal}$ (e) $SRGAN_{Ledig}$ (\cite{Srgan}) ; (f) $SRCNN$ (\cite{SRCNN}); (g) SR-RF (\cite{SR-RF}); (h) SSR (\cite{Jog}). }
\label{fig:RetSR2}
\end{figure*}

Figure~\ref{fig:RetSR2} shows super resolution results for retinal fundus images. The image has been downsampled by a factor of $16$ each along rows and columns, and the original image is obtained using different methods. $SRGAN_{Sal}$ shows better reconstruction results (Fig.~\ref{fig:RetSR2} (d) )than most competing methods. However the reconstruction is a bit blurry for small retinal vessels using $SRGAN_{Sal}$ as indicated by the yellow arrow. This defect is overcome using the proposed $P-SRGAN$ architecture (Fig.~\ref{fig:RetSR2} (c) ), thus justifying the use of progressive GANs over saliency maps for image super resolution.

\begin{table}
\begin{tabular}{|c|c|c|c|c|}
\hline
{} & \multicolumn {4}{|c|}{Scaling factor = $4$} \\  \hline
{} & {SSIM}  & {PSNR} & {S3} & {$p$} \\ 
{} & {} & {(dB)} & {} & {} \\ \hline
{$P-SRGAN$} & {0.91} & {46.1} & {0.85} & {-} \\ \hline
{$SRGAN_{Sal}$} & 0.88 & {44.3}  & 0.82 & $<0.001$  \\ \hline
{$P-SRGAN_{MSE}$} & {0.77} & {37.0} & {0.68} & {-} \\ \hline
{$SRGAN_{Ledig}$ \cite{Srgan}} & {0.76} & 36.7 & 0.66 & $<0.001$ \\  \hline
{$SRCNN$ \cite{SRCNN}}  & {0.74} & 34.3 & 0.61 & $<0.0009$ \\  \hline
{SR-RF \cite{SR-RF}}   & {0.72} & 30.3 & 0.59 & $<0.0009$  \\  \hline
{SSR \cite{Jog}}  & {0.68} & 27.2 & 0.57   & $<0.001$ \\  \hline
{} & \multicolumn {4}{|c|}{Scaling factor = $8$} \\  \hline
{} & {SSIM}  & {PSNR} & {S3} & {$p$} \\ 
{} & {} & {(dB)} & {} & {} \\ \hline
{$P-SRGAN$} & {0.81} & {40.3} & {0.75} & {-} \\ \hline
{$SRGAN_{Sal}$} & 0.77 & 36.2 & 0.70 & $<0.001$ \\ \hline
{$P-SRGAN_{MSE}$} & {0.72} & {31.9} & {0.62} & {$<0.001$} \\ \hline
{$SRGAN_{Ledig}$ \cite{Srgan}}  & 0.71 & 31.4 & 0.60 & $<0.001$   \\  \hline
{$SRCNN$ \cite{SRCNN}}  &  0.65 & 28.4 & 0.57 &$<0.001$   \\  \hline
{SR-RF \cite{SR-RF}}   & 0.63 & 25.6 & 0.55 & $<0.001$  \\  \hline
{SSR \cite{Jog}}  &  0.60 & 22.3 & 0.51 & $<0.001$   \\  \hline
{} & \multicolumn {4}{|c|}{Scaling factor = $16$} \\  \hline
{} & {SSIM}  & {PSNR} & {S3} & {$p$} \\ 
{} & {} & {(dB)} & {} & {} \\ \hline
{$P-SRGAN$} & {0.77} & {36.8} & {0.68} & {-} \\ \hline
{$SRGAN_{Sal}$} & 0.71 & 31.8 & 0.60 & $<0.001$ \\ \hline
{$P-SRGAN_{MSE}$} & {0.66} & {30.1} & {0.56} & {$<0.001$} \\ \hline
{$SRGAN_{Ledig}$ \cite{Srgan}}  & 0.66 & 29.4 & 0.55 & $<0.001$   \\  \hline
{$SRCNN$ \cite{SRCNN}}  &  0.60 & 23.2 & 0.52 &$<0.001$   \\  \hline
{SR-RF \cite{SR-RF}}   & 0.57 & 21.5 & 0.50 & $<0.001$  \\  \hline
{SSR \cite{Jog}}  &  0.52 & 19.4 & 0.46 & $<0.001$   \\  \hline
{} & \multicolumn {4}{|c|}{Scaling factor = $32$} \\  \hline
{} & {SSIM}  & {PSNR} & {S3} & {$p$} \\ 
{} & {} & {(dB)} & {} & {} \\ \hline
{$P-SRGAN$} & {0.72} & {31.9} & {0.61} & {-} \\ \hline
{$SRGAN_{Sal}$} & 0.67 & 28.6 & 0.57 & $<0.001$ \\ \hline
{$P-SRGAN_{MSE}$} & {0.62} & {24.2} & {0.54} & {$<0.001$} \\ \hline
{$SRGAN_{Ledig}$ \cite{Srgan}}  & 0.61 & 23.7 & 0.53 & $<0.001$   \\  \hline
{$SRCNN$ \cite{SRCNN}}  &  0.55 & 19.4 & 0.49 &$<0.001$   \\  \hline
{SR-RF \cite{SR-RF}}   & 0.52 & 17.4 & 0.47 & $<0.001$  \\  \hline
{SSR \cite{Jog}}  &  0.49 & 15.8 & 0.45 & $<0.001$   \\  \hline
\end{tabular}
\caption{Comparative results of different methods for image super resolution.}
\label{tab:ISR_res1}
\end{table}

%
%

\subsection{Robustness to Noise}
 
In this section we show results for cases when noise was added to the LR images and the high resolution image was reconstructed by using $P-SRGAN$, $SRGAN_{Sal}$ and $SRGAN_{Ledig}$. We added the following types of noise: 1) Gaussian noise with zero mean and standard deviation varying from $[0.001,.01]$ in steps of $0.001$; 2) salt and pepper noise of density varying from $[0.01,0.05]$ increasing in steps of $0.01$; 3) speckle noise of variance from $[0.01,0.05]$ increasing in steps of $0.01$ to the LR image (subsampled to factor $8$) and attempted to obtain the original high resolution image through super resolution methods. Table~\ref{tab:ISR_noise} summarizes the average performance for each noise type. We observe that as noise intensity increases the performance worsens. Although the overall performance is not as good compared to the noise free images, the results for gaussian noise case are still within acceptable limits for our proposed method.

\begin{table}
\begin{tabular}{|c|c|c|c|c|}
\hline
{} & \multicolumn {4}{|c|}{Gaussian Noise} \\  \hline
{} & {SSIM}  & {PSNR} & {S3} & {$p$} \\ 
{} & {} & {(dB)} & {} & {} \\ \hline
{$P-SRGAN$} & {0.81$\pm$0.04} & {39.5$\pm$2.1} & {0.78$\pm$0.05} & {$<0.002$} \\ \hline
{$SRGAN_{Sal}$} & 0.73$\pm$0.06 & 36.1$\pm$4.2 & 0.66$\pm$0.06 & $<0.001$ \\ \hline
{$SRGAN_{Ledig}$ \cite{Srgan}} & {0.72$\pm$0.06} & 32.4$\pm$4.9 & 0.61$\pm$0.06 & $<0.001$ \\  \hline
{} & \multicolumn {4}{|c|}{Salt and Pepper Noise} \\  \hline
{} & {SSIM}  & {PSNR} & {S3} & {$p$} \\ 
{} & {} & {(dB)} & {} & {} \\ \hline
{$P-SRGAN$} & {0.74$\pm$0.07} & {35.8$\pm$4.5} & {0.68$\pm$0.07} & {$<0.001$} \\ \hline
{$SRGAN_{Sal}$} & 0.69$\pm$0.1 & 31.4$\pm$5.7 & 0.63$\pm$0.1 & $<0.001$ \\ \hline
{$SRGAN_{Ledig}$ \cite{Srgan}}  & 0.60$\pm$0.11 & 23.4$\pm$7.6 & 0.56$\pm$0.12 & $<0.001$   \\  \hline
{} & \multicolumn {4}{|c|}{Speckle Noise} \\  \hline
{} & {SSIM}  & {PSNR} & {S3} & {$p$} \\ 
{} & {} & {(dB)} & {} & {} \\ \hline
{$P-SRGAN$} & {0.72$\pm$0.09} & {33.5$\pm$8.2} & {0.63$\pm$0.09} & {$<0.001$} \\ \hline
{$SRGAN_{Sal}$} & 0.67$\pm$0.11 & 28.8$\pm$9.8 & 0.58$\pm$0.10 & $<0.001$ \\ \hline
{$SRGAN_{Ledig}$ \cite{Srgan}}  & 0.62$\pm$0.13 & 2.4$\pm$10.2 & 0.51$\pm$0.11 & $<0.001$   \\  \hline
\end{tabular}
\caption{Comparative results for image super resolution at subsampling factor $8$ and added noise - mean and variance values are provided.}
\label{tab:ISR_noise}
\end{table}

\subsection{Retinal Blood Vessel Segmentation Results}

A major application of super resolution of retinal images is better image analysis as in accurate landmark segmentation and pathology detection. 
We present retinal vessel segmentation results on our dataset \cite{kaggle}. $40$ images from the dataset had the retinal vessels manually annotated by an expert. This is comparable to other public datasets where a maximum number of $40$ images had the retinal vessels manually delineated.
We take the original images (HR images) and downsample them by factors of $4-16$ to get LR images. All the different algorithms were used to generate SR images of the original scale. The super resolution models were trained on the $5000$ training images described before. The SR images are then used to train a U-Net based method for retinal vessel segmentation \cite{Ret_Unet}. We use the method in \cite{Ret_Unet} since it shows the best results on the DRIVE dataset.

The idea of developing ISR algorithms is to demonstrate that they generate good quality high resolution images from a low resolution.
A possible use case is in tele-ophthalmology where images acquisition is done by a low quality camera. Even if a high resolution device is used, the image still needs to be compressed from high to low resolution for transmission over a network. In this case, the quality of the image received by the clinician is often not ideal for visualisation and interpretation.
%
In such cases we would expect the generated SR images would do a better job at tasks like segmentation and pathology detection. To evaluate our hypothesis we adopt the following evaluation procedure for the SR images:
\begin{enumerate}
\item As described before we use the SR images generated from downsampled images to train different UNet segmentation networks. Additionally we also generate segmentation results using the original HR images to train another UNet architecture. Segmentation results from this network give an upper bound on the best possible super resolution performance.

\item To ascertain whether the super resolved images lead to improved image analysis over LR images, we also train UNets on the downsampled LR images by using the correspondingly downsampled ground truths. The results on these images (for scales $4,8,16$) act as a baseline to compare super resolution algorithms. If ISR is successful then segmentation performance of SR images should be better than the results from these LR images, and reasonably close to segmentation results obtained from original HR images.  
\end{enumerate}

Table~\ref{tab:SrRes_vasc}  shows the average segmentation performance using training images from different ISR algorithms. In Table~\ref{tab:SrRes_vasc} $HR$ denotes the fact that the network was trained on the reference HR images (of dimension $1024\times1024$). $LR_{n}$ denotes the training and testing was done on subsampled images at scale $n$ (i.e., image dimension $1024/n\times1024/n$). Results of the other methods denote the segmentations on low resolution images super resolved to dimension $1024\times1024$.

Retinal vessel performance of different methods was evaluated using accuracy ($Acc$) and sensitivity ($Sen$).  
The segmentation results for different scaling factors gives interesting insight into the different super resolution algorithms. As expected, the HR images give the best possible results since they represent the original images while other images have some level of information loss due to subsampling and super resolution. Our proposed method gives better segmentation performance than $LR_n$ which clearly indicates that our ISR algorithm generates better quality images than the low resolution versions. 

Interestingly methods such as $SSR$ and $SR-RF$ perform only slightly better than $LR_n$. A major factor is the super resolution algorithm. These two methods use traditional approaches based on defining a mathematical model. The other methods that perform better than $LR_n$ however use neural network models for ISR. Amongst the neural network based models the ones using generator networks perform better than $SRCNN$ because the generative models learn to generate better images based on the similarity with the higher resolution images. This clearly  indicates that the generator models do a much better job in learning the underlying the image representation, while mathematical models only learn a limited aspect of the image. 

Comparing between the generative models our proposed method does a better job than \cite{Srgan} because of the progressive stages used to generate the images. From the second stage onward our use of triplet loss improves image quality by making it as similar as possible to the HR image. Hence we obtain improved image quality.  Note that results of $P-SRGAN_{MSE}$ are very similar to that of  $SRGAN_{Ledig}$. This clearly indicates the improvements brought about by using the triplet loss function

\begin{figure*}[t]
\begin{tabular}{ccccc}
\includegraphics[height=2.7cm,width=2.6cm]{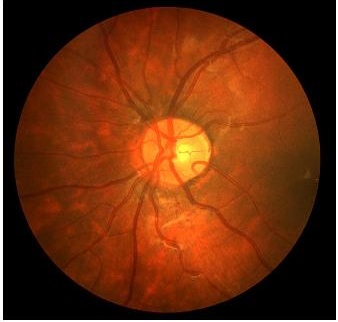} &
\includegraphics[height=2.7cm,width=2.6cm]{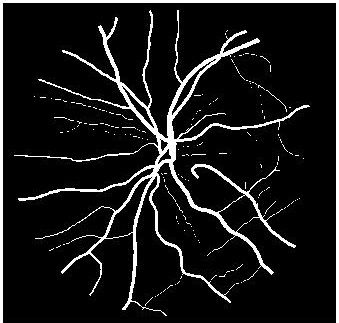} &
\includegraphics[height=2.7cm,width=2.6cm]{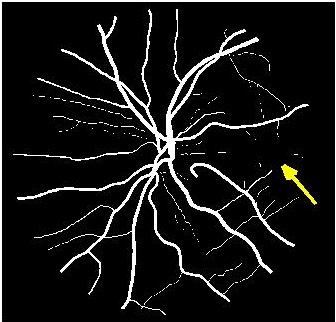} &
\includegraphics[height=2.7cm,width=2.6cm]{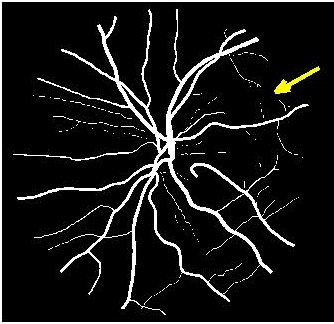} &
\includegraphics[height=2.7cm,width=2.6cm]{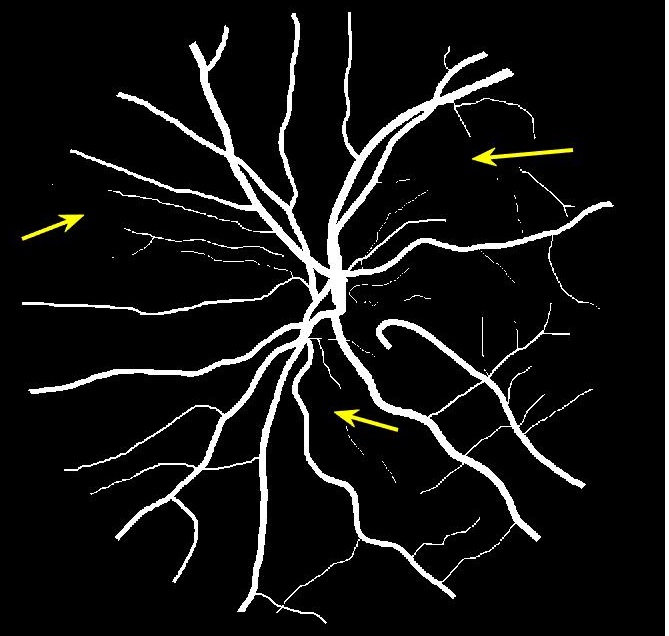} \\

(a) & (b) & (c) & (d) & (e) \\
\includegraphics[height=2.7cm,width=2.6cm]{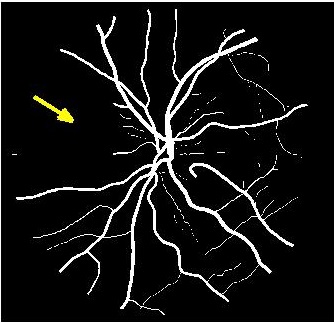} &
\includegraphics[height=2.7cm,width=2.6cm]{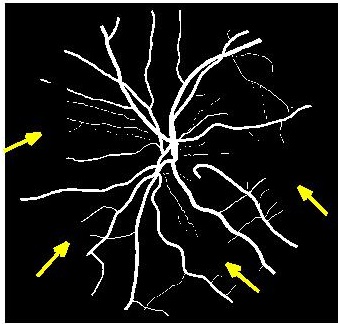} &
\includegraphics[height=2.7cm,width=2.6cm]{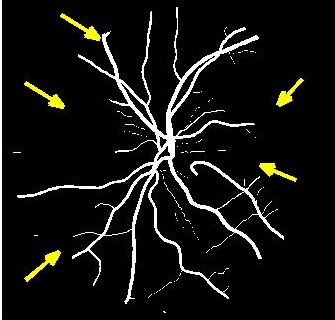} &
\includegraphics[height=2.7cm,width=2.6cm]{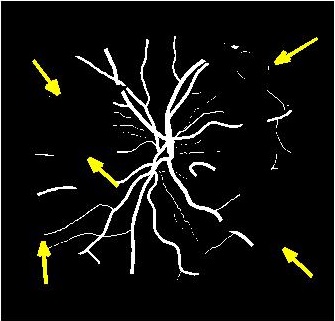} &
\includegraphics[height=2.7cm,width=2.6cm]{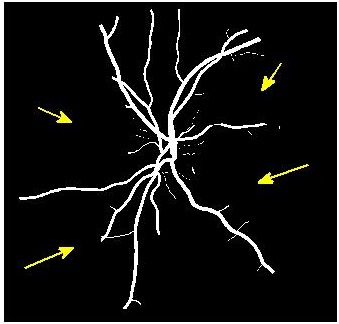} \\

(f) & (g) & (h) & (i) & (j)\\ 
\end{tabular}
\caption{Example results for retinal vasculature segmentation; (a) retinal image; (b) manual ground truth mask; results obtained when training on (c) orginal HR images; SR images from: (d)  $P-SRGAN$; (e) $SRGAN_{Sal}$; (f) SR images by $SRGAN_{Ledig}$ (\cite{Srgan}) ; (g) SR images using $SRCNN$ (\cite{SRCNN}); (h) SR images using SR-RF (\cite{SR-RF}); (i) SR images using SSR (\cite{Jog}); (j) LR image of scale factor $8$. Yellow arrows highlight regions of inaccurate segmentation.}
\label{fig:res1}
\end{figure*}

Figure~\ref{fig:res1} (a) shows an example retinal image followed by its ground truth manual segmentation in Figure~\ref{fig:res1} (b). Figure~\ref{fig:res1} (c) shows segmentation result for scaling factor $8$ when using the original HR images to train the U-Net followed by the results when trained on the super resolved images generated by $P-SRGAN$, (Figure~\ref{fig:res1} (d)), $SRGAN_{Ledig}$ (Figure~\ref{fig:res1} (e)),  $SRCNN$ (Figure~\ref{fig:res1} (f)), SR-RF (Figure~\ref{fig:res1} (g)), SSR (Figure~\ref{fig:res1} (h)) and $LR_8$ (Figure~\ref{fig:res1} (i)). Obviously the results from $P-SRGAN$ provide results most similar to those of HR images. This is also validated by the quantitative results in Table~\ref{tab:SrRes_vasc}. The areas where different methods are unable to obtain accurate segmentation are highlighted by yellow arrows. Due to poor quality of super resolved images most of the methods do not segment the finer vasculature structures, while SSR and SR-RF are unable to segment some of the major arteries. Importantly, our method performs much better than the low resolution image ($LR_8$) which performs poorly due to low resolution.

\begin{table}
\begin{tabular}{|c|c|c|c|c|c|c|}
\hline
{} & \multicolumn {2}{|c|}{Scale Factor = 4} & \multicolumn {2}{|c|}{Scale Factor = 8} & \multicolumn {2}{|c|}{Scale Factor = 16} \\  \hline
{} & {Acc} & {Sen}  & {Acc} & {Sen} &{Acc} & {Sen} \\ \hline
HR & 0.98  & 0.91  & 0.98  & 0.90  & 0.97 &  0.84 \\  \hline
{$P-SRGAN$} & {0.96} & {0.89}  & 0.89 & 0.85 & 0.84 & 0.80 \\ \hline
{$SRGAN_{Sal}$} & {0.93} & {0.86}  & 0.86 & 0.81 & 0.82 & 0.76 \\ \hline
{$P-SRGAN_{MSE}$} & {0.92} & {0.84}  & 0.85 & 0.79 & 0.79 & 0.74 \\ \hline
$SRGAN_{Ledig}$ (\cite{Srgan})  & 0.91  & 0.83  & 0.85  & 0.80  & 0.79  & 0.74 \\  \hline
$SRCNN$ (\cite{SRCNN})  & 0.88 & 0.81 & 0.83 &  0.78 &  0.76 &  0.73 \\  \hline
SR-RF (\cite{SR-RF}) & 0.85  & 0.77  & 0.80  & 0.73  & 0.74 &  0.70 \\  \hline
SSR (\cite{Jog})  & 0.83 & 0.75 & 0.77 & 0.69 & 0.71 & 0.68 \\  \hline
$LR_n$  & 0.81 & 0.72 & 0.75 & 0.67 & 0.70 & 0.66 \\  \hline
\end{tabular}
\caption{Comparative vasculature segmentation results of different super resolution methods at different scale factors.}
\label{tab:SrRes_vasc}
\end{table}

\subsection{Results for Retinal Microaneurysm Detection}

We also present results for microaneurysm (MA) detection. $300$ images from \cite{kaggle} are annotated by outlining the boundaries of different occurences of MAs. 
MA's are very difficult to detect as they cover only a very small area in a high dimensional retinal image. The intuition behind applying image super resolution to MA detection is if the resolution of the patch containing MA's can be increased then detection accuracy by machine learning algorithms would improve.

 We design a pipeline where a U-Net architecture is trained to segment these MAs. For each image we extract $256\times256$ patches covering the annotated MAs. These are the original HR images from which LR images of size $128\times128$, $64\times64$ and $32\times32$ are obtained. The different super resolution algorithms are used to generate $256\times256$ patches from the LR patches leading to scaling factors of $2,4,8$. For every generated SR image of different scaling factors, we train U-Nets on images from the different super resolution algorithms. 

We also train a U-Net segmentation framework with the original $256\times256$ HR images which gives the lower bound of performance error. Similar to vessel segmentation we also generate segmentation results on the LR images, $LR_{n}$, that act as the baseline metric.
For our experiments we employ $5-$fold cross validation. Multiple patches extracted from the entire are transformed by rotation and translation such that the final training set size was $100,000$ patches. Note that UNet is a patch based segmentation framework and hence we use patches for its training.

After the segmentations are obtained for each set of test images, we calculate the following metrics: sensitivity ($Sen$), Specificity ($Spe$) and area under curve (AUC). These values are summarized in Table~\ref{tab:MA_res} with respect to scaling factors $4,8$ since the results are very similar for scaling factor $2$. 
It is obvious from the results that  super resolution is highly effective for scaling factor $4$,  and to a lesser extent for factor $8$. We can infer that super resolved images by all the competing methods for scaling factor $8$ changes the information content in the image to the extent that their performance is significantly worse than the HR images. Despite these constraints our algorithm gives the best performance amongst all methods, and most importantly it performs better than $LR_n$. It is important to note that for pathologies, super resolution for scale factor higher than $8$ is not feasible since the LR image dimensions are too small.

\begin{table}
\begin{tabular}{|c|c|c|c|c|}
\hline
 \multicolumn {5}{|c|}{Scaling factor = $4$} \\  \hline
{} & {HR} & {$P-SRGAN$} & {$SRGAN$}  & {$SRCNN$}\\ 
{} & {} & {} & {\cite{Srgan}}  & {\cite{SRCNN}}\\ \hline
{$Sen$} & {0.83} & 0.81 & {0.77} & {0.73}   \\ \hline
{$Spe$} & {0.80} & {0.77} & {0.72} & 0.69 \\  \hline
{AUC} & {0.94}  & {0.92} & {0.86}  & 0.83   \\  \hline
{$p-$}   &  {-} & {$<0.001$}  & $<0.0023$ & $<0.0001$ \\  \hline
{} & {$SRGAN_{Sal}$} & {$SR-RF$} & {$SSR$} & {$LR_4$}\\ 
{} & {\cite{Mahapatra_MICCAI17}} & {\cite{SR-RF}} & {\cite{Jog}} & {}\\ \hline
{$Sen$} & 0.79  & 0.71 &  {0.68} & {0.67} \\ \hline
{$Spe$} & 0.74 & 0.65 & 0.63  &  {0.63}\\  \hline
{AUC} & 0.88 & 0.80 & 0.78  &{0.77}    \\  \hline
{$p-$}   &  {$<0.002$} & $<0.003$ & $<0.009$ & $<0.009$ \\  \hline
\multicolumn {5}{|c|}{Scaling factor = $8$} \\  \hline
{} & {HR} & {$P-SRGAN$} & {$SRGAN$}  & {$SRCNN$}\\ 
{} & {} & {} & {\cite{Srgan}}  & {\cite{SRCNN}}\\ \hline
{$Sen$} & {} &  {0.80} & 0.76 & {0.72}  \\ \hline
{$Spe$} & {} & 0.74 & 0.69 & 0.66   \\  \hline
{AUC} & {}  & 0.86 & 0.81 & 0.77   \\  \hline
{$p-$}   &  {-} & $<0.00065$ & $<0.0002$ & $<0.0001$ \\  \hline
{} & {$SRGAN_{Sal}$} & {$SR-RF$} & {$SSR$} & {$LR_4$}\\ 
{} & {\cite{Mahapatra_MICCAI17}} & {\cite{SR-RF}} & {\cite{Jog}} & {}\\ \hline
{$Sen$} & {0.77} & {0.68}  & 0.67 & {0.65} \\ \hline
{$Spe$} & {0.70} & 0.64 & 0.63 & 0.61  \\  \hline
{AUC} & {0.83}  & 0.74 & 0.70 & 0.69  \\  \hline
{$p-$}   &  {$<0.004$} & $<0.0087$ & $<0.0001$ & $<0.0001$\\  \hline
\end{tabular}
\caption{Comparative MA detection results of different image super resolution methods.}
\label{tab:MA_res}
\end{table}

\begin{figure}[t]
\begin{tabular}{cccc}
\includegraphics[height=2.9cm,width=2.7cm]{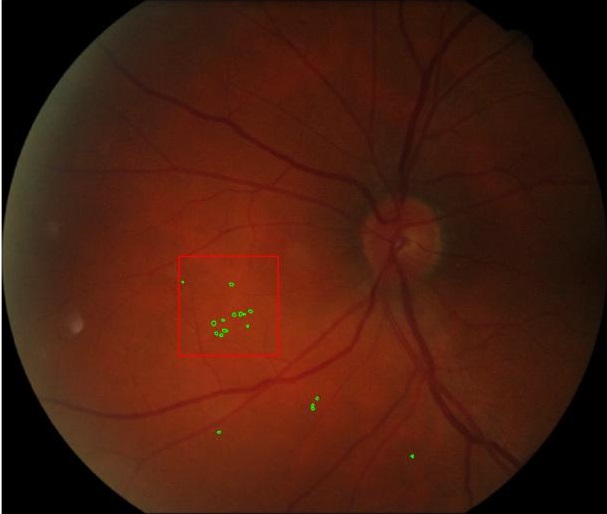} &
\includegraphics[height=2.9cm,width=2.7cm]{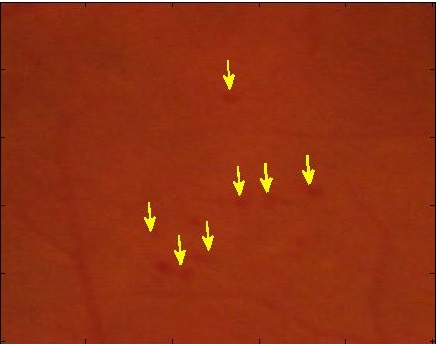} &
\includegraphics[height=2.9cm,width=2.7cm]{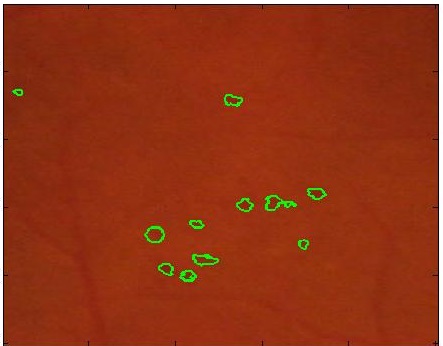} &
\includegraphics[height=2.9cm,width=2.7cm]{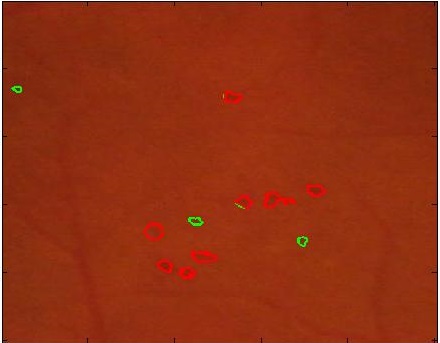} \\
(a) & (b) & (c) & (d) \\
\includegraphics[height=2.9cm,width=2.7cm]{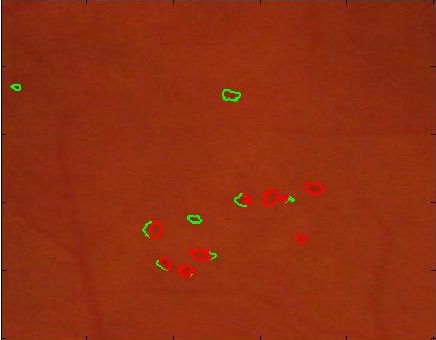} &
\includegraphics[height=2.9cm,width=2.7cm]{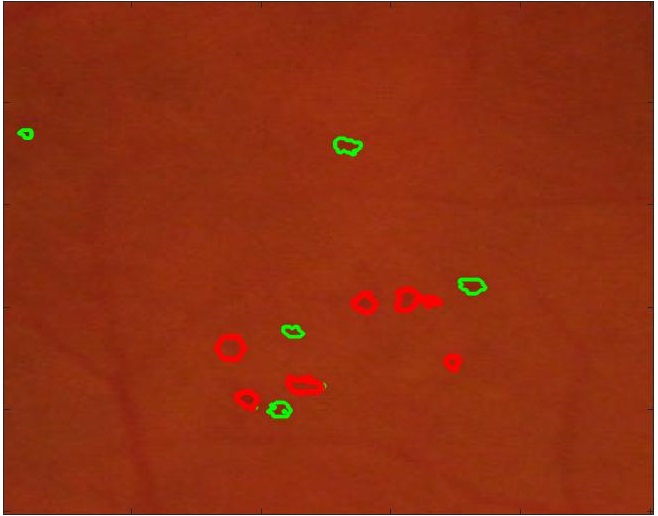} &
\includegraphics[height=2.9cm,width=2.7cm]{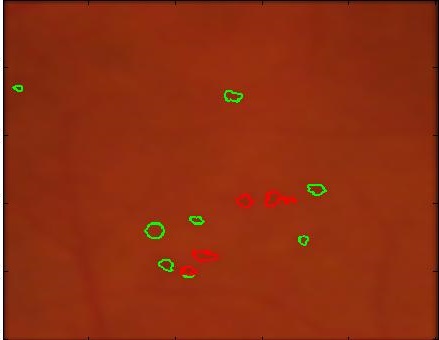} &
\includegraphics[height=2.9cm,width=2.7cm]{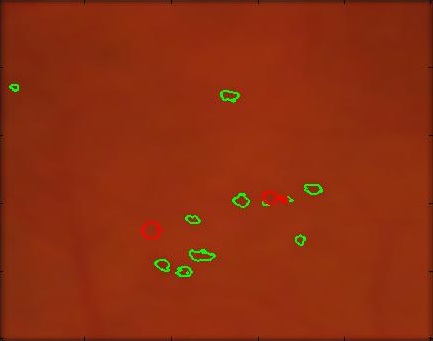} \\
(e) & (f) & (g) & (h) \\ 
\end{tabular}
\caption{Example results for microaneurysm segmentation using super resolved images obtained using different methods. (a) original image with microaneurysm region outline by red square; (b) cropped version of original image with yellow arrows denoting MA locations; (c) manual annotations of MA regions; super resolution and segmentation results obtained when trained on (d) original HR images; (e) SR images by $P-SRGAN$; (f) SR images by $SRGAN_{Sal}$ (g) SR images by $SRGAN_{Ledig}$ (\cite{Srgan}); (h) SR images using $SRCNN$ (\cite{SRCNN}). Green contours are ground truth annotations and algorithm segmentations are denoted by red contours.}
\label{fig:MASeg1}
\end{figure}

Figure~\ref{fig:MASeg1} shows results of MA segmentation using the images generated by different super resolution methods. Figure~\ref{fig:MASeg1} (a) shows the original full sized image with the regions having majority of the MAs outlined by a red square. Figure~\ref{fig:MASeg1} (b) shows the cropped image region with yellow arrows identifying location of MAs. Figure~\ref{fig:MASeg1} (c) shows manually drawn contours around the MAs. Figure~\ref{fig:MASeg1} (d) shows the segmentation results obtained using the original HR images as part of the training and test set. The ground truth manual contours are shown in green while the segmentations obtained using the U-Net algorithm is shown in red. The performance on the HR images is an indication of the minimum error (or best possible segmentation performance). Figures~\ref{fig:MASeg1} (e)-(h) show, respectively, the super resolved images obtained by $P-SRGAN$, $SRGAN_{Sal}$, $SRGAN_{Ledig}$ and $SRCNN$ along with the super imposed segmentation results. It is quite obvious that the results obtained using $P-SRGAN$ are the best. It is interesting to note that the SR images obtained using $SRGAN_{Ledig}$ and $SRCNN$ lead to blurred edges of the blood vessels and MAs, although in the case of $SRGAN_{Sal}$ the SR images are not blurred. The other two methods, $SR-RF$ and $SSR$ result in such poor quality images that the MAs are not even detected for this particular example. 

The MA segmentation results clearly show that if we generate a high resolution image from a low resolution image using our method and then run any segmentation or detection algorithm on the output, the results will be very close to what we would get if we had acquired the image in the high resolution setting. This has immense significance in clinical applications that require detection of small pathologies in a high resolution image. The clinician can simply select the suspect area and our algorithm can generate a high resolution image that can be used for subsequent analysis.

\subsection{Results on MRI Dataset}

Although retinal fundus images have a high resolution during acquisition time, ISR is still relevant to detect small pathologies and landmarks. ISR's impact can be further judged by its performance on MRI since MR images have low-resolution during acquisition time. Furthermore, the anatomies to be detected in MRI cover an even smaller area. Hence there is greater need for reliable ISR algorithms which facilitate better performance in detection or segmentation of pathologies. We demonstrate the relevance of ISR on cardiac MR images.

\subsubsection{Cardiac MR Data}
We used images from the Sunnybrook Cardiac Dataset \cite{Sunnybrook}. The 
images used in this evaluation study are cine steady state
free precession (SSFP) MR short axis (SAX) images. All images were obtained
on a 1.5T GE Signa MR scanner during 10 to 15s breath
holds with a temporal resolution of 20 cardiac phases over
the heart cycle (thickness=8-10mm, FOV= $320 \times 320$mm,
matrix=$256 \times 256$). The experimental procedures were approved by
the Institutional Review Board. The ground truth of the inner and outer
boundaries of the left ventricle (endocardium and epicardium
respectively) is manually delineated by the clinical experts
from Sunnybrook Health Sciences Centre. The training set has images from $15$  patients, and the number of slices for each patient varied from $6-12$

We train our ISR network from scratch. Each volume slice is treated as a separate image and transformed by random rotation and translation. We rotate the images between $[0,180^{\circ}]$ in steps of $9^{\circ}$. For each rotation we translate the image by $[0,50]$ pixels in steps of $2$. Thus on an average we get $20*25=500$ transformation for each image. Thus an average of $10$ slices per image gives a total of $15\times10\times500=75,000$  cardiac MR images. We show results on ISR for scale factor $4,8$. Results for ISR are summarized in Tables~\ref{tab:cardiacISR}. Following the approach for retinal landmark and pathology segmentation, we also show results for cardiac left ventricle segmentation (Table~\ref{tab:LVSeg}). For each segmentation approach we employ UNets as the segmentation framework and show results for different super-resolution methods as well as the low-resolution images ($LR_n$). Dice metric values for segmentation accuracy are shown in each case. Similar to retinal pathology segmentation we extract a $256\times256$ patch covering the pathology and apply super-resolution for scale factors $4,8$.

\begin{figure*}[t]
\begin{tabular}{cccc}
\includegraphics[height=3.2cm,width=3.2cm]{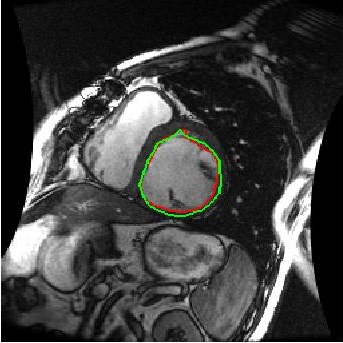} &
\includegraphics[height=3.2cm,width=3.2cm]{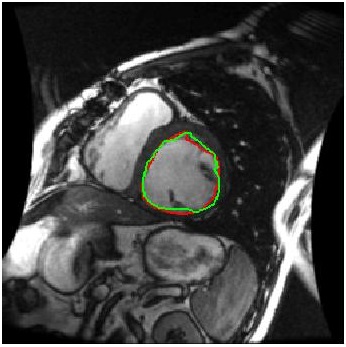} &
\includegraphics[height=3.2cm,width=3.2cm]{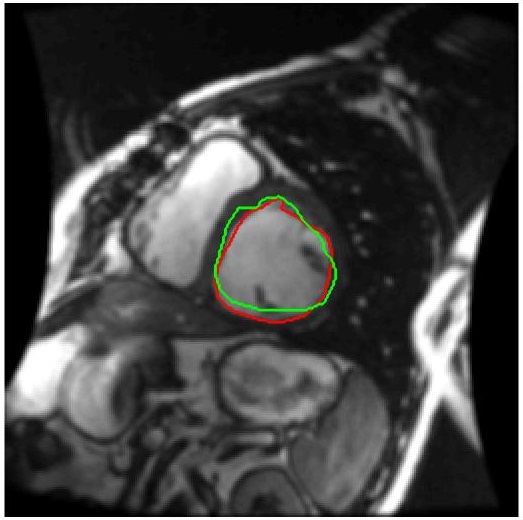} &
\includegraphics[height=3.2cm,width=3.2cm]{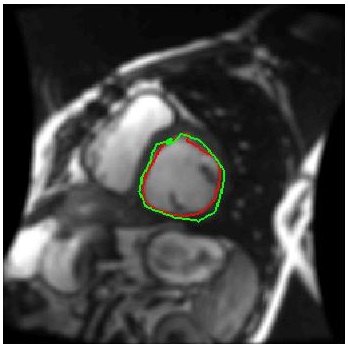} \\
(a) & (b) & (c) & (d) \\

\includegraphics[height=3.2cm,width=3.2cm]{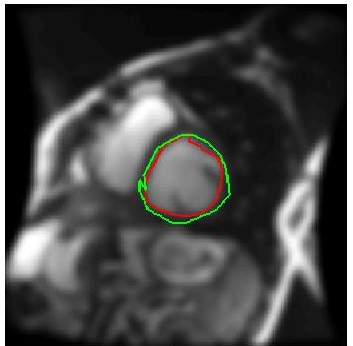} &
\includegraphics[height=3.2cm,width=3.2cm]{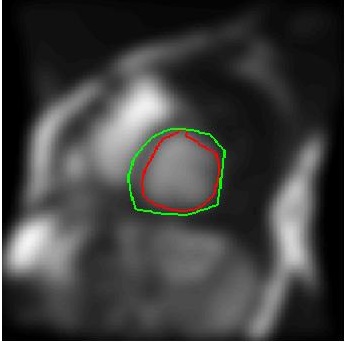} &
\includegraphics[height=3.2cm,width=3.2cm]{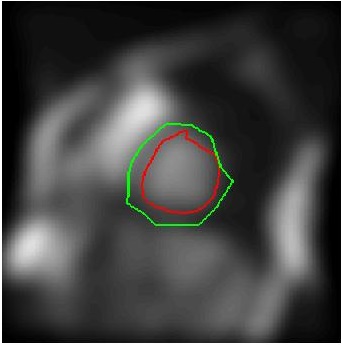} &
\includegraphics[height=3.2cm,width=3.2cm]{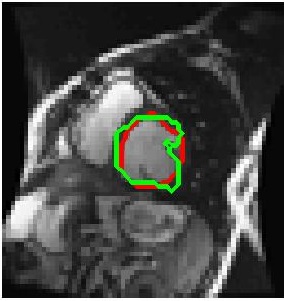} \\

(e) & (f) & (g) & (h)\\ 
\end{tabular}
\caption{Example results for cardiac left ventricle segmentation. Results obtained when training on (a) orginal HR images; (b) SR images by $P-SRGAN$; (c) SR images by $SRGAN_{Sal}$; (d) SR images by $SRGAN_{Ledig}$ (\cite{Srgan}) ; (e) SR images using $SRCNN$ (\cite{SRCNN}); (f) SR images using SR-RF (\cite{SR-RF}); (g) SR images using SSR (\cite{Jog}); and (h) LR images subsampled by scale $4$. Red contour is ground truth while green contour is the algorithm segmentation.}
\label{fig:CardSRSeg}
\end{figure*}

Figure~\ref{fig:CardSRSeg} shows results for segmenting the cardiac LV from MRI. For  each case we present results on the original HR images, SR images obtained by each of the $5$ methods being compared and also when using the LR  images (scale factor $4$). It is quite obvious that the LR images are very fuzzy and don't  give accurate information on the anatomical boundaries. On the other hand the SR images from our method can predict a highly accurate reconstruction of the actual image. Other ISR methods show some degree of blur in the SR images. It is remarkable that deep neural network based methods are able to reconstruct original high quality images despite limited information in LR images. This is possible because of the ability of the generator networks to learn the relation between HR and LR images.

\begin{table}
\begin{tabular}{|c|c|c|c|c|}
\hline
{} & \multicolumn {4}{|c|}{Scaling factor = $4$} \\  \hline
{} & {SSIM}  & {PSNR} & {S3} & {$p$} \\ 
{} & {} & {(dB)} & {} & {} \\ \hline
{$P-SRGAN$} & {0.83} & {43.7} & {0.81} & {-} \\ \hline
{$SRGAN_{Sal}$} & 0.78 & {38.4}  & 0.75 & $<0.001$ \\ \hline
{$SRGAN_{Ledig}$ (\cite{Srgan})} & {0.70} & 34.1 & 0.70 & $<0.001$ \\  \hline
{$SRCNN$ (\cite{SRCNN})}  & {0.69} & 32.7 & 0.65 & $<0.0009$ \\  \hline
{SR-RF (\cite{SR-RF})}   & {0.66} & 31.2 & 0.63 & $<0.0009$  \\  \hline
{SSR (\cite{Jog})}  & {0.63} & 28.3 & 0.61   & $<0.001$ \\  \hline
{} & \multicolumn {4}{|c|}{Scaling factor = $8$} \\  \hline
{} & {SSIM}  & {PSNR} & {S3} & {$p$} \\ 
{} & {} & {(dB)} & {} & {} \\ \hline
{$P-SRGAN$} & {0.79} & {41.3} & {0.78} & {-} \\ \hline
{$SRGAN_{Sal}$} &  0.74 & 37.0 & 0.72 & $<0.001$ \\ \hline
{$SRGAN_{Ledig}$ (\cite{Srgan})}  & 0.67 & 31.7 & 0.66 & $<0.001$   \\  \hline
{$SRCNN$ (\cite{SRCNN})}  &  0.64 & 29.8 & 0.61 &$<0.001$   \\  \hline
{SR-RF (\cite{SR-RF})}   & 0.62 & 28.2 & 0.59 & $<0.001$  \\  \hline
{SSR (\cite{Jog})}  &  0.59 & 26.1 & 0.58 & $<0.001$   \\  \hline
\end{tabular}
\caption{Comparative results of different methods for cardiac image super-resolution.}
\label{tab:cardiacISR}
\end{table}


\begin{table}
\begin{tabular}{|c|c|c|c|c|c|c|c|c|}
\hline
{} & \multicolumn {2}{|c|}{$HR$} & \multicolumn {2}{|c|}{$P-GAN$} & \multicolumn {2}{|c|}{$SRGAN_{Ledig}$} & \multicolumn {2}{|c|}{$SRGAN_{Sal}$} \\  
{} & \multicolumn {2}{|c|}{} & \multicolumn {2}{|c|}{} & \multicolumn {2}{|c|}{\cite{Srgan}} & \multicolumn {2}{|c|}{\cite{Mahapatra_MICCAI17}} \\  \hline
{Scale (n)} & {DM} & {HD}  & {DM} & {HD}& {DM} & {HD} & {DM} & {HD} \\ \hline
{$4$} & {91.3} & {6.9} & {90.1} & {7.2} & {86.4} & {9.0} & {88.2} & {8.1} \\ \hline 
{$8$} & {87.4} & {8.8} & {86.8} & {9.2} & {83.0} & {10.6} & {84.7} & {9.8} \\ \hline 
{} & \multicolumn {2}{|c|}{$SRCNN$} & \multicolumn {2}{|c|}{$SR-RF$} & \multicolumn {2}{|c|}{$SSR$} & \multicolumn {2}{|c|}{$LR_n$} \\  
{} & \multicolumn {2}{|c|}{\cite{SRCNN}} & \multicolumn {2}{|c|}{\cite{SR-RF}} & \multicolumn {2}{|c|}{\cite{Jog}} & \multicolumn {2}{|c|}{} \\  \hline
{Scale (n)} & {DM} & {HD} & {DM} & {HD} & {DM} & {HD} & {DM} & {HD} \\ \hline
{$4$} & {84.8} & {9.7} & {82.3} & {11.0} & {81.1} & {11.4} & {83.4} & {10.4} \\ \hline 
{$8$} & {81.1} & {11.5} & {79.8} & {11.9} & {78.7} & {12.4} & {80.7} & {11.6} \\ \hline 
\end{tabular}
\caption{Comparative cardiac LV segmentation results with different super-resolution methods at different scale factors.}
\label{tab:LVSeg}
\end{table}

\section{Conclusion}
\label{sec:concl}

We have proposed a novel method for super-resolution of different medical images based on progressive generative adversarial networks that combines triplet loss with the conventional MSE and CNN loss in a multi stage architecture. The core novelty of our method lies in incorporating the triplet loss into the cost function of PGANs. The triplet loss ensures that super resolved images produced in one stage undergo quality improvement for the next stage. Consequently our method is able to preserve the high quality of images for high scaling factors (greater than $8$). The super resolved images obtained using our method are of much better quality than the original GAN framework that uses MSE and CNN loss. The quality improvement of our super resolved images is evident from the image quality metrics. Our proposed method's superior performance is also demonstrated in the case of retinal vessel segmentation and microaneurysm detection, as well MRI super-resolution and cardiac LV segmentation. 

The resulting super resolved images can be used to increase the size and resolution of low dimensional images, and different image analysis algorithms can be applied to the super resolved images. As demonstrated through our results, the SR images obtained by our method performs better than the low-resolution images from which the super resolved images are obtained. The results from using our super resolved images will be very close to that when using the original high-resolution images for segmentation or pathology detection.


\section{References}
\bibliographystyle{elsarticle-num}
\bibliography{ISR_GAN_Ref}

\begin{thebibliography}{10}
\expandafter\ifx\csname url\endcsname\relax
  \def\url#1{\texttt{#1}}\fi
\expandafter\ifx\csname urlprefix\endcsname\relax\def\urlprefix{URL }\fi
\expandafter\ifx\csname href\endcsname\relax
  \def\href#1#2{#2} \def\path#1{#1}\fi

\bibitem{Jog7}
J.~Manjon, P.~Coupe, A.~Buades, V.~Fonov, D.~Collins, M.~Robles, Non-local mri
  upsampling., Med. Image Anal. 14~(6) (2010) 784--792.

\bibitem{Jog8}
F.~Rousseau, Brain hallucination, in: ECCV, 2008, pp. 497--508.

\bibitem{Jog5}
E.~Konukoglu, A.~van~der Kouwe, M.~Sabuncu, B.~Fischl., Example-based
  restoration of high-resolution magnetic resonance image acquisitions, in: In
  Proc MICCAI, 2013, pp. 131--138.

\bibitem{Jog9}
A.~Rueda, N.~Malpica, E.~Romero., Single-image super-resolution of brain mr
  images using overcomplete dictionaries., Med. Image Anal. 17~(1) (2013)
  113--132.

\bibitem{Oktay3}
K.~Bhatia, A.~Price, W.~Shi, J.~Hajnal, D.~Rueckert., Super-resolution
  reconstruction of cardiac mri using coupled dictionary learning, in: In Proc
  ISBI, 2014, pp. 947--950.

\bibitem{TannoMICCAI16}
R.~Tanno, A.~Ghosh, F.~Grussu, E.~Kaden, A.~Criminisi, D.~Alexander., Bayesian
  image quality transfer, in: In Proc MICCAI, 2016, pp. 265--173.

\bibitem{Oktay5}
C.~Dong, Y.~Deng, C.~Loy, X.~Tang., Compression artifacts reduction by a deep
  convolutional network, in: In Proc IEEE ICCV, 2015, pp. 576--584.

\bibitem{Jog}
A.~Jog, A.~aron Carass, J.~Prince., Self super-resolution for magnetic
  resonance images, in: In Proc MICCAI, 2016, pp. 553--560.

\bibitem{Oktay}
O.~Oktay, et. al., Multi-input cardiac image super-resolution using
  convolutional neural networks, in: In Proc MICCAI, 2016, pp. 246--254.

\bibitem{Srgan}
C.~Ledig, et. al., Photo-realistic single image super-resolution using a
  generative adversarial network, CoRR abs/1609.04802.

\bibitem{Mahapatra_MICCAI17}
D.~Mahapatra, B.~Bozorgtabar, S.~Hewavitharanage, R.~Garnavi, Image super
  resolution using generative adversarial networks and local saliency maps for
  retinal image analysis, in: MICCAI, 2017, pp. 382--390.

\bibitem{Srgan21}
I.~Goodfellow, J.~Pouget-Abadiey, M.~Mirza, B.~Xu, D.~Warde-Farley, S.~Ozairz,
  A.~Courville, Y.~Bengio., Generative adversarial nets, in: Proc. NIPS., 2014,
  pp. 2672--2680.

\bibitem{MahapatraGAN_ISBI18}
D.~Mahapatra, B.~Antony, S.~Sedai, R.~Garnavi, Deformable medical image
  registration using generative adversarial networks, in: In Proc. IEEE ISBI,
  2018, pp. 1449--1453.

\bibitem{MahapatraAL_MICCAI18}
D.~Mahapatra, S.~Bozorgtabar, J.-P. Thiran, M.~Reyes, Efficient active learning
  for image classification and segmentation using a sample selection and
  conditional generative adversarial network, in: In Proc. MICCAI (2), 2018,
  pp. 580--588.

\bibitem{CondGANs}
P.~Isola, J.~Zhu, T.~Zhou, A.~Efros, Image-to-image translation with
  conditional adversarial networks, in: CVPR, 2017.

\bibitem{CyclicGANs}
J.~Zhu, T.park, P.~Isola, A.~Efros, Unpaired image-to-image translation using
  cycle-consistent adversarial networks, in: arXiv preprint arXiv:1703.10593,
  2017.

\bibitem{Srgan27}
K.~He, X.~Zhang, S.~Ren, J.~Sun, Deep residual learning for image recognition,
  in: Proc. CVPR., 2016, pp. 770--778.

\bibitem{VGG}
K.~Simonyan, A.~Zisserman., Very deep convolutional networks for large-scale
  image recognition, CoRR abs/1409.1556.

\bibitem{TripletLoss}
F.~Schroff, D.~Kalenichenko, J.~Philbin, Facenet: A unified embedding for face
  recognition and clustering, in: The IEEE Conference on Computer Vision and
  Pattern Recognition (CVPR), 2015, pp. 815--823.

\bibitem{kaggle}
https://www.eyepacs.com.

\bibitem{SRCNN}
C.~Dong, C.~Loy, K.~He, X.~Tang, Image super-resolution using deep
  convolutional networks., IEEE Trans. Patt. Anal. Mach. Intell. 38~(2) (2016)
  295--307.

\bibitem{SR-RF}
S.~Schulter, C.~Leistner, H.~Bischof., Fast and accurate image upscaling with
  super-resolution forests, in: In Proc CVPR, 2015, pp. 3791--3799.

\bibitem{SSIM}
Z.~Wang, et. al., Image quality assessment: from error visibility to structural
  similarity., IEEE Trans. Imag. Proc. 13~(4) (2004) 600--612.

\bibitem{S3Metric}
C.~Vu, T.~Phan, D.~Chandler., S3: {A} spectral and spatial measure of local
  perceived sharpness in natural images., IEEE Trans. Imag. Proc. 21~(3) (2012)
  934--945.

\bibitem{Ret_Unet}
https://github.com/orobix/retina-unet.

\bibitem{Sunnybrook}
P.~Radau, Y.~Lu, K.~Connelly, G.~Paul, et.~al ., valuation framework for
  algorithms segmenting short axis cardiac {MRI}, in: The MIDAS Journal-Cardiac
  MR Left Ventricle Segmentation Challenge, 2009.

\end{thebibliography}

\end{document}